\documentclass{egpubl}
\usepackage{eg2026}
 
\ConferencePaper        
\CGFStandardLicense

\usepackage[T1]{fontenc}
\usepackage{dfadobe}  

\usepackage{cite}  
\BibtexOrBiblatex
\electronicVersion
\PrintedOrElectronic
\ifpdf \usepackage[pdftex]{graphicx} \pdfcompresslevel=9
\else \usepackage[dvips]{graphicx} \fi

\usepackage{egweblnk} 

\usepackage{amssymb}
\usepackage{amsmath}
\usepackage{algpseudocode}
\usepackage{algorithm}
\usepackage{multirow}

\usepackage{hyperref}
\usepackage[capitalise]{cleveref}

\usepackage{pgfplots}
\usepackage{subcaption}
\usepackage{placeins}

\usepackage{booktabs}
\usepackage{xcolor}
\usepackage{colortbl}
\usepackage{makecell}

\title[TLC-Plan: A Two-Level Codebook Based Network for End-to-End Vector Floorplan Generation]%
{TLC-Plan: A Two-Level Codebook Based Network for End-to-End Vector Floorplan Generation}

\author[B. Xiong et al.]
{\parbox{\textwidth}{\centering
Biao~Xiong$^{1}$\orcid{0000-0001-8106-3799},
Zhen~Peng$^{1}$,
Ping~Wang$^{1}$,
Qiegen~Liu$^{2}$\orcid{0000-0003-4717-2283},
and Xian~Zhong$^{1,}$\thanks{Corresponding author: zhongx@whut.edu.cn. This work was supported by the National Natural Science Foundation of China (Grant No. 62271361) and the Hubei Provincial Key Research and Development Program (Grant No. 2024BAB039).}\orcid{0000-0002-5242-0467}} \\
{\parbox{\textwidth}{\centering
$^{1}$ Hubei Key Laboratory of Transportation Internet of Things, School of Computer Science and Artificial Intelligence, \\ Wuhan University of Technology, Wuhan 430070, China \\
$^{2}$ School of Information Engineering, Nanchang University, Nanchang 330031, China
}}
}

\begin{document}


\maketitle
\begin{abstract}
Automated floorplan generation aims to improve design quality, architectural efficiency, and sustainability by jointly modeling global spatial organization and precise geometric detail. However, existing approaches operate in raster space and rely on post hoc vectorization, which introduces structural inconsistencies and hinders end-to-end learning. Motivated by compositional spatial reasoning, we propose TLC-Plan, a hierarchical generative model that directly synthesizes vector floorplans from input boundaries, aligning with human architectural workflows based on modular and reusable patterns. TLC-Plan employs a two-level VQ-VAE to encode global layouts as semantically labeled room bounding boxes and to refine local geometries using polygon-level codes. This hierarchy is unified in a CodeTree representation, while an autoregressive transformer samples codes conditioned on the boundary to generate diverse and topologically valid designs, without requiring explicit room topology or dimensional priors. Extensive experiments show state-of-the-art performance on \textsc{RPLAN} dataset (FID = 1.84, MSE = 2.06) and leading results on \textsc{LIFULL} dataset. The proposed framework advances constraint-aware and scalable vector floorplan generation for real-world architectural applications. Source code and trained models are released at https://github.com/rosolose/TLC-PLAN.

\ccsdesc[500]{Computing methodologies~Machine learning}
\ccsdesc[300]{Theory of computation~Computational geometry}
\ccsdesc[300]{Computing methodologies~Graphics systems and interfaces}

\printccsdesc 
\end{abstract} 
\section{Introduction}

Automatic floorplan generation is a central problem in artificial intelligence and architectural design, with broad impact on computer-aided design, game-level creation, interior planning, construction automation, and sustainable urban development~\cite{meselhy2025review,wang2023survey}. A floorplan specifies the arrangement of rooms and functional spaces within a building, typically represented by walls on a two-dimensional plane~\cite{cogo2024survey}. Effective synthesis must simultaneously satisfy architectural constraints such as adjacency and alignment, while producing diverse and plausible designs that support downstream applications, including site-scale layout generation~\cite{wang2024automated}, BIM model construction~\cite{zeng2025automated}, and daylight performance optimization~\cite{hu2025prediction}.

\begin{figure}[!t]
	\centering
	\includegraphics[width = \linewidth]{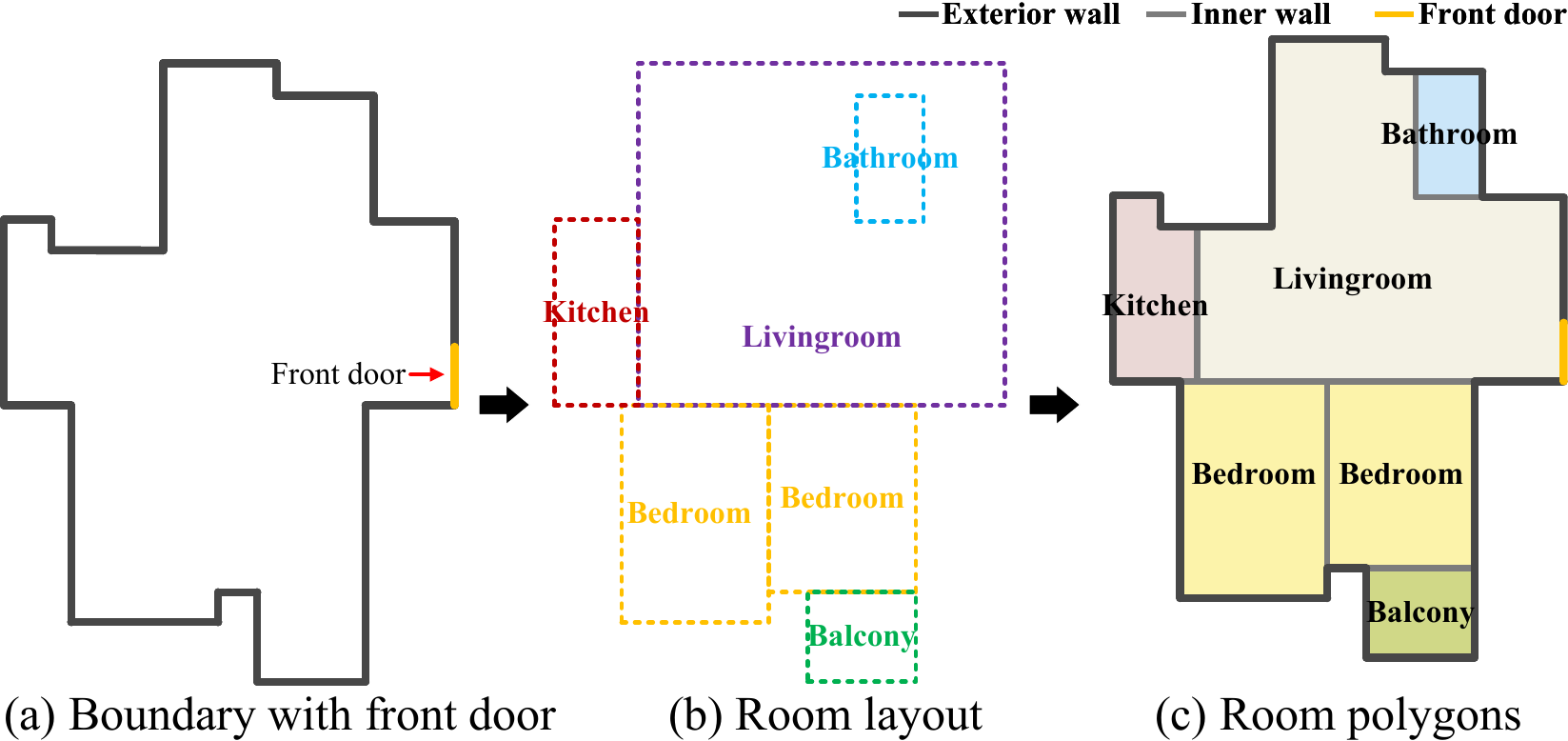}
	\caption{\textbf{TLC-Plan directly generates vector floorplans end-to-end in two stages:} (a) an input boundary with front-door location, (b) a layout-level codebook predicts semantically labeled room bounding boxes, and (c) a polygon-level codebook refines each box into a detailed room polygon. The resulting floorplan is geometrically aligned and CAD-ready, requiring no post-processing.}
	\label{fig:teaser}
\end{figure}

Early approaches relied on procedural rules or expert-guided optimization~\cite{cogo2024survey}, using stochastic search~\cite{merrell2010computer} or portal graphs~\cite{bao2013generating} to construct residential layouts. More recent convolutional network-based methods learn layout distributions from real floorplans, but typically generate rasterized representations followed by vectorization~\cite{he2022iplan,zhang2024maskplan}. Although convolutional and adversarial networks can capture global structure, rasterization introduces discretization artifacts, leading to misaligned walls and corners during non-differentiable vectorization and hindering end-to-end training~\cite{liu2025computer}. Vector-based methods, including FloorplanGAN’s hybrid generator-discriminator~\cite{luo2022floorplangan}, Graph2Plan’s graph networks with predefined adjacency~\cite{hu2020graph2plan}, and HouseDiffusion’s denoising framework~\cite{shabani2023housediffusion}, alleviate some of these issues but depend on handcrafted graphs or other manual inputs, which limits compositional generalization.

We argue that effective floorplan generation requires disentangled representations of global layout and local geometry, analogous to macro-level planning in urban design and micro-level detailing in building shapes~\cite{liu2025cage,wang2025local}. Based on this insight, we introduce TLC-Plan, an end-to-end framework for structured vector floorplan synthesis (see \Cref{fig:teaser}). TLC-Plan employs a two-level vector-quantized variational autoencoder (VQ-VAE) to encode floorplans into discrete latent codes: top-level codes model room layouts, while bottom-level codes refine precise room polygons. Given only an input boundary, an autoregressive transformer samples a CodeTree that combines these codes and decodes it into a topologically and geometrically aligned, computer-aided design (CAD)-ready floorplan. Experiments on \textsc{RPLAN}~\cite{wu2019data} and \textsc{LIFULL}~\cite{lifull2016} datasets show that TLC-Plan achieves state-of-the-art performance, including an FID of 1.84 and an MSE of 2.06 on \textsc{RPLAN}, while exhibiting superior geometric fidelity and strong generalization across diverse layout distributions. Our main contributions are threefold:

\begin{itemize}
	\item We introduce TLC-Plan, a novel hierarchical VQ-VAE framework for direct end-to-end vector floorplan synthesis from boundaries, eliminating rasterization artifacts and advancing structured generative modeling in spatial AI.
	\item We propose a two-level discrete latent space that separates global layouts from local geometries, capturing architectural hierarchies without requiring room topology or dimensional priors, and enabling compositional generalization aligned with design principles.
	\item Extensive evaluations on \textsc{RPLAN} and \textsc{LIFULL} datasets demonstrate that TLC-Plan achieves state-of-the-art performance in coherence, precision, and diversity, outperforming existing baselines and supporting scalable applications in automated architecture.
\end{itemize}

\section{Related Work}\label{sec:related}

\subsection{Raster-based Floorplan Generation}

Early learning-based approaches model floorplans as multi-channel raster images with pixel-wise room labels. \textsc{RPLAN}~\cite{wu2019data} rasterizes layouts and applies heuristic vector tracing, which often introduces blur and structural inconsistencies. Graph2Plan~\cite{hu2020graph2plan} predicts room bounding boxes from a predefined adjacency graph, but its separate alignment module can fail on irregular boundaries. WallPlan~\cite{sun2022wallplan} improves semantic understanding through a wall-graph decoder, yet enforces geometric consistency only as a post-processing step. PlanNet~\cite{fu2023plannet} retrieves similar rasterized layouts instead of learning high-level design priors, limiting its generative flexibility.

Subsequent methods incorporate user interactivity and semantic conditioning. ActFloorGAN~\cite{wang2021actfloor} synthesizes room layouts guided by human activity maps. iPLAN~\cite{he2022iplan} supports interactive editing via cascaded masks, but repeated raster operations accumulate geometric drift. MaskPLAN~\cite{zhang2024maskplan} employs dynamic masked autoencoders for flexible editing, yet remains constrained by pixel grids. Overall, raster-based pipelines depend on post hoc vectorization and struggle to achieve CAD-level geometric precision.

\subsection{Vector-based Floorplan Generation}

Vector-based methods represent layouts with explicit geometric or topological constraints. MIQP~\cite{wu2018miqp} formulates floorplan synthesis as a mixed-integer quadratic program that decomposes layouts into rectangles under size, position, and adjacency constraints. HouseGAN and HouseGAN++~\cite{nauata2020house,nauata2021house} generate axis-aligned room boxes using relational GANs, but the rectangular assumption precludes curved walls and diagonal structures. Constraint-graph-based approaches~\cite{para2021generative,liu2022end} predict nodes and edges before solving an external optimization, which breaks end-to-end learning. G2Plan~\cite{bisht2022transforming} and BubbleFormer~\cite{sun2023bubbleformer} generate bubble diagrams from boundaries but stop short of producing full vector floorplans.

Structured and disentangled representation learning has also been explored in other domains~\cite{yu2025sparse}. 
More recent work adopts diffusion models to improve geometric fidelity. HouseDiffusion~\cite{shabani2023housediffusion}, Cons2Plan~\cite{hong2024cons2plan}, and GSDiff~\cite{hu2025gsdiff} denoise corner or graph representations, but still rely on bubble diagrams and multi-stage pipelines. DiffPlanner~\cite{wang2025eliminating} iteratively refines layouts, yet separates topology from geometry. In contrast, TLC-Plan learns a unified two-level CodeTree that jointly captures global layout and room-level geometry, enabling direct end-to-end vector floorplan generation without external graph supervision.

\subsection{LLM-Guided Layout and Scene Design}

Recent advances in large language models (LLMs) enable semantic control over spatial design through natural language, with applications spanning mechanical design~\cite{wang2025cad}, BIM modeling~\cite{du2024text2bim}, visual storytelling~\cite{yang2025storyllava}, and mobility analysis~\cite{jiawei2024large}. HouseLLM~\cite{zong2024housellm} translates user prompts into bubble diagrams via chain-of-thought reasoning and diffusion, but degrades under long or ambiguous inputs. I-Design and Holodeck~\cite{ccelen2025idesign,yang2024holodeck} support 3D scene arrangement, yet depend on fixed geometry templates. LayoutVLM~\cite{sun2024layoutvlm} optimizes image-space layouts but cannot produce vector graphics.

Despite their semantic flexibility, most LLM-driven systems lack the geometric precision required for CAD-faithful floorplans. Mamba-CAD~\cite{li2025mamba} models long CAD sequences through self-supervised learning, while CAD2Program and RECAD~\cite{wang20252d,li2025revisiting} reconstruct 3D parametric models from 2D sketches using vision-language and diffusion frameworks. By contrast, TLC-Plan introduces a discrete, geometry-aware code space that preserves vector accuracy, bridging semantic reasoning and precise floorplan synthesis, and providing a foundation for future LLM-driven architectural design.


\section{Proposed Method}
\label{sec:method}



We present \textbf{TLC-Plan}, a two-stage generative framework that directly synthesizes high-quality vector floorplans from building boundaries. Inspired by architectural design principles and hierarchical modeling in HNC-CAD~\cite{xu2023hierarchical}, TLC-Plan decomposes floorplan generation into two stages: global layout planning and room-level geometric refinement. Unlike HNC-CAD, which considers only geometric information, TLC-Plan explicitly incorporates semantic room types at the layout level and structural cues at the polygon level, enabling functional spatial reasoning. This hierarchy is realized through parallel VQ-VAE networks that learn a two-level codebook, which is then synthesized by an autoregressive transformer conditioned solely on the input boundary.


\begin{figure}[!t]
	\centering
	\includegraphics[width = \linewidth]{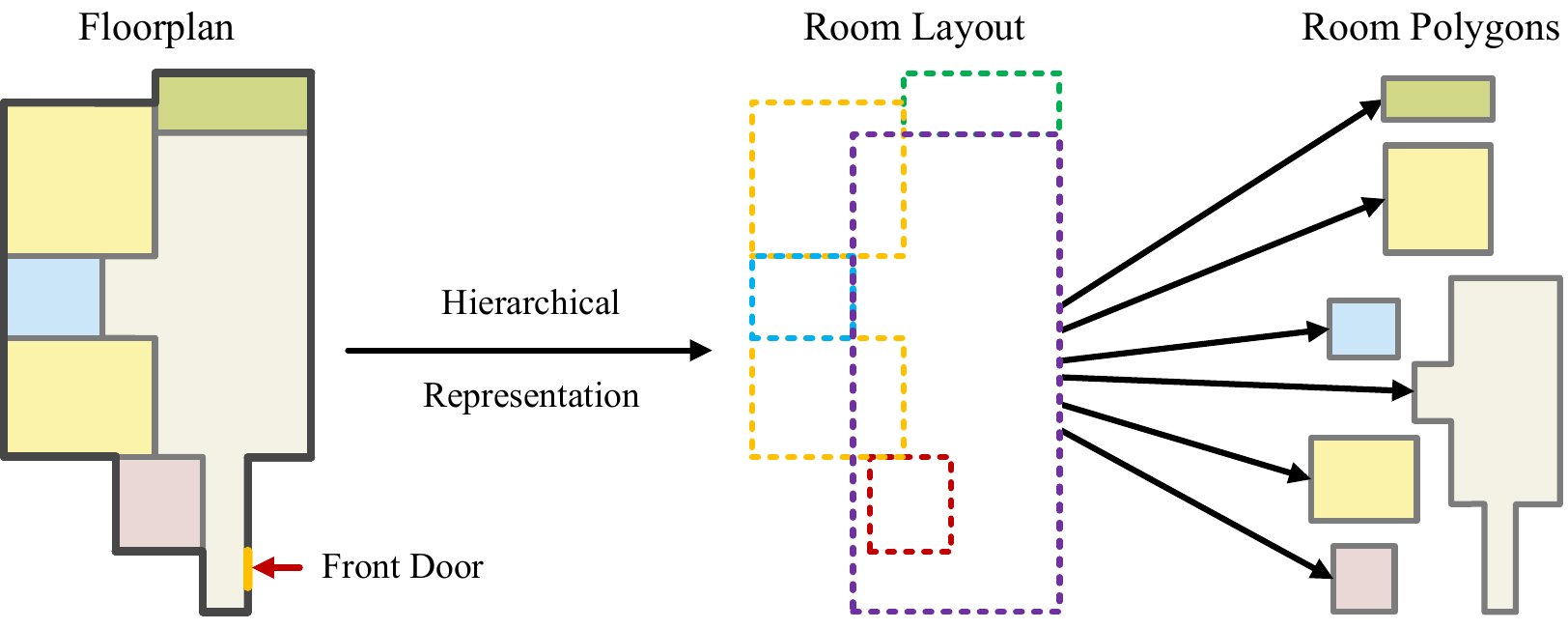}
	\caption{\textbf{Hierarchical two-level representation:} the layout level captures global room bounding boxes, while the polygon level refines them into detailed shapes, enabling structured codebook learning.}
	\label{fig:two_level_rep}
\end{figure}

\subsection{Hierarchical Floorplan Representation}


Vectorized floorplans provide precise geometric and semantic descriptions required by CAD and architectural applications. TLC-Plan adopts a two-level hierarchical representation~\cite{yang2025storyllava} that decomposes a floorplan into a high-level room layout and low-level polygon geometries, as shown in \Cref{fig:two_level_rep}.

\subsubsection{Room Layout}

The global layout $L$ is represented as an ordered sequence of room descriptors $(x_i, y_i, w_i, h_i, c_i)$ for $i = 1, \dots, M$, where $(x_i, y_i)$ denote the bottom-left coordinates, $(w_i, h_i)$ the room dimensions, and $c_i$ the room type:
\begin{align}\label{eq:layout}
	L = \left\{\left(x_i, y_i, w_i, h_i, c_i \right) \right\}_{i = 1}^{M}.
\end{align}
Rooms are ordered by placing the living room first, followed by the remaining rooms sorted by increasing $x$ and then $y$ coordinates.

\subsubsection{Room Polygons}

Each room polygon $P_i$ is defined as a clockwise-ordered list of vertices:
\begin{align}\label{eq:polygon}
	P_i = \left\{\left(x_1, y_1 \right), \left(x_2, y_2 \right), \dots, \left(x_n, y_n \right) \right\}.
\end{align}
This vertex list delineates the geometric contour of the room polygon. The front door, if present, is encoded within the polygon by reordering the vertex sequence such that the two vertices corresponding to the door appear first, while preserving the overall clockwise orientation of the polygon.

\subsubsection{Floorplan CodeTree}

A complete floorplan is represented by its global layout $L$ and the set of room polygons $\{P_i\}_{i=1}^M$:
\begin{align}
	F = \left[L, P_1, P_2, \dots, P_M \right].
\end{align}
To enable compact representation and efficient generation, we discretize $L$ and each $P_i$ into codebook indices $c_L \in \mathcal{B}_L$ and $c_{P_i} \in \mathcal{B}_P$, yielding a hierarchical CodeTree:
\begin{align}
	C = \left[c_L, c_{P_1}, c_{P_2}, \dots, c_{P_M} \right].
\end{align}
This discrete sequence jointly captures global spatial configuration and detailed room geometry, facilitating modular learning and autoregressive sampling.



\begin{figure*}[!t]
	\centering
	\includegraphics[width = \linewidth]{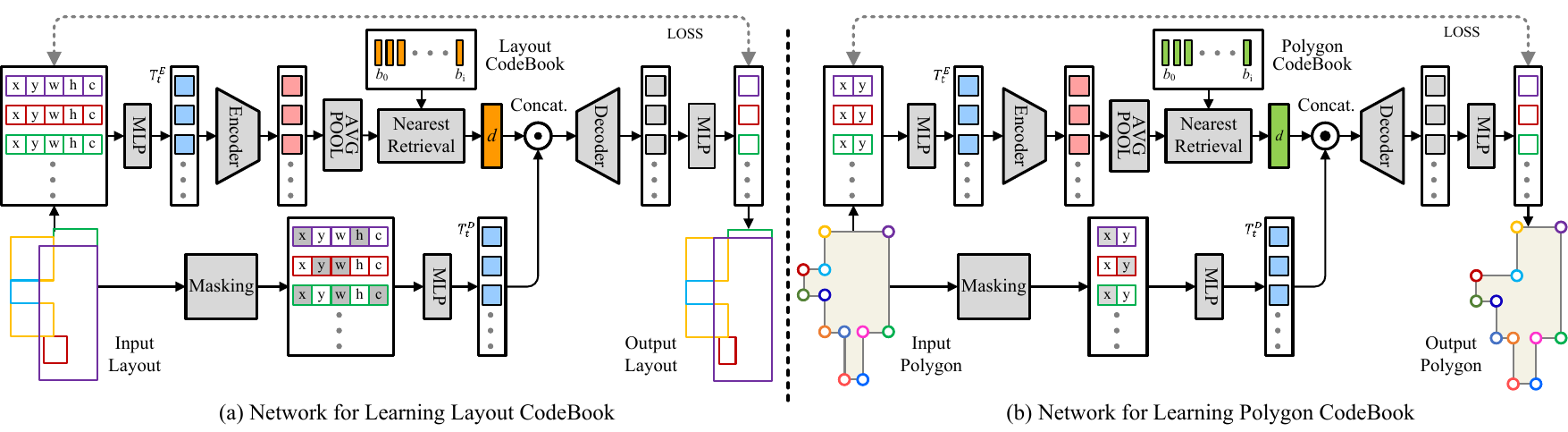}
	\caption{\textbf{Network architecture for codebook learning:} layout-level and polygon-level models share the same architecture but differ in input tokenization. The encoder extracts sequence features, which are quantized via a learned codebook, and the decoder reconstructs masked inputs from quantized codes to capture reusable design patterns.}
	\label{fig:codebook}
\end{figure*}

\subsection{Two-Level Codebook Learning}\label{sec:codebook}


To learn modular and reusable design abstractions, TLC-Plan adopts a dual VQ-VAE framework~\cite{oord2018neuraldiscreterepresentationlearning,razavi2019generating}: one encoder-decoder pair for room layouts (see \Cref{fig:codebook}(a)) and another for room polygons (see \Cref{fig:codebook}(b)). The two models share the same architecture but use different tokenization and embedding strategies.

\subsubsection{Room Embedding and Encoding}

Each layout tuple $(x, y, w, h, c)$ is discretized into 6-bit values and embedded into 32-D vectors. The resulting 320-D room vector is projected to a 256-D token using a two-layer MLP with positional encoding:
\begin{align}
\label{eq:encoder}
	T_t^E = \mathrm{MLP} \left(W_g x_t \| W_g y_t \| W_g w_t \| W_g h_t \| W_c c_t \right) + \gamma_t,
\end{align}
where $\|$ denotes concatenation, $W_g$ embeds spatial components $(x,y,w,h)$, $W_c$ embeds the room type, and $\gamma_t$ is the positional encoding. A Transformer encoder processes the sequence $\{T_t^E\}$, and average pooling yields the layout feature $\overline{E}(T^E)$.

\subsubsection{Vector Quantization}

The encoded feature is quantized via nearest-neighbor search:
\begin{align}
	d = b_k, \quad k = \arg \min_i \left\|\overline{E} \left(T^E \right) - b_i \right\|^2,
\end{align}
where $\{b_i\}$ are codebook entries.


\subsubsection{Masked Reconstruction}


To encourage abstraction learning, we randomly mask 30-70\% of the input embeddings and require the decoder to reconstruct the missing values from the quantized codes. This masking strategy prevents direct memorization and forces the codebooks to capture essential structural patterns.

\subsubsection{Codebook Update}


TLC-Plan updates both the layout and polygon codebooks using an Exponential Moving Average (EMA) scheme, avoiding backpropagation through discrete code indices\cite{hou2016squared}. During training, the encoder maps input tokens to continuous embeddings, which are quantized by assigning each embedding to its nearest codeword. For each codeword, EMA updates are applied by accumulating the assigned embeddings and their counts, and then updating the codeword as the normalized moving average of these embeddings. This process gradually adapts the codebook to the data distribution while maintaining training stability. The masked reconstruction objective encourages codewords to represent reusable design patterns, and the commitment loss ensures encoder outputs remain close to their assigned codewords.


\subsubsection{Loss Function}

The overall objective combines reconstruction, codebook, and commitment losses:
\begin{align}
\begin{split}
	\mathcal{L} = & \sum_t \mathrm{EMD} \left(D \left(d, \left\{T_t^D \right\} \right), \mathbb{1}_{T_t} \right) \\
	+ & \left\|\mathrm{sg} \left[\overline{E} \left(T^E \right) \right] - d \right\|^2
	+ \beta \left\|\overline{E} \left(T^E \right) - \mathrm{sg} \left[d \right] \right\|^2,
\end{split}
\end{align}
with $\beta = 0.25$ and $\mathrm{sg}[\cdot]$ the stop-gradient operator. We use the squared Earth Mover's Distance (EMD)~\cite{hou2017squaredearthmoversdistancebased} for reconstruction:
\begin{align}
	\mathrm{EMD} \left(D \left(d, \left\{T_t^D \right\} \right), \mathbb{1}_{T_t} \right)
	= \frac{1}{C} \sum_{c=1}^C \left( \sum_{i=1}^c p_{t,i} - \sum_{i=1}^c q_{t,i} \right)^2,
\end{align}
where $\boldsymbol{p}_t$ and $\boldsymbol{q}_t$ denote the predicted distribution $D(d,\{T_t^D\})$ and the target distribution $\mathbb{1}_{T_t}$, respectively.

\subsubsection{Polygon-Level Model}

The polygon-level VQ-VAE mirrors the layout-level model but operates on vertex sequences instead of bounding boxes, enabling precise room-level geometric modeling.

\subsection{Vector Floorplan Generation}

TLC-Plan synthesizes complete vector floorplans from input boundaries by autoregressively predicting a discrete \emph{CodeTree} that jointly encodes global layout structure and detailed room geometry. During training, ground-truth CodeTrees provide supervision. At inference time, the model samples a CodeTree conditioned on boundary features and decodes it into a coherent vector floorplan.

\begin{figure*}[!t]
	\centering
	\includegraphics[width = \linewidth]{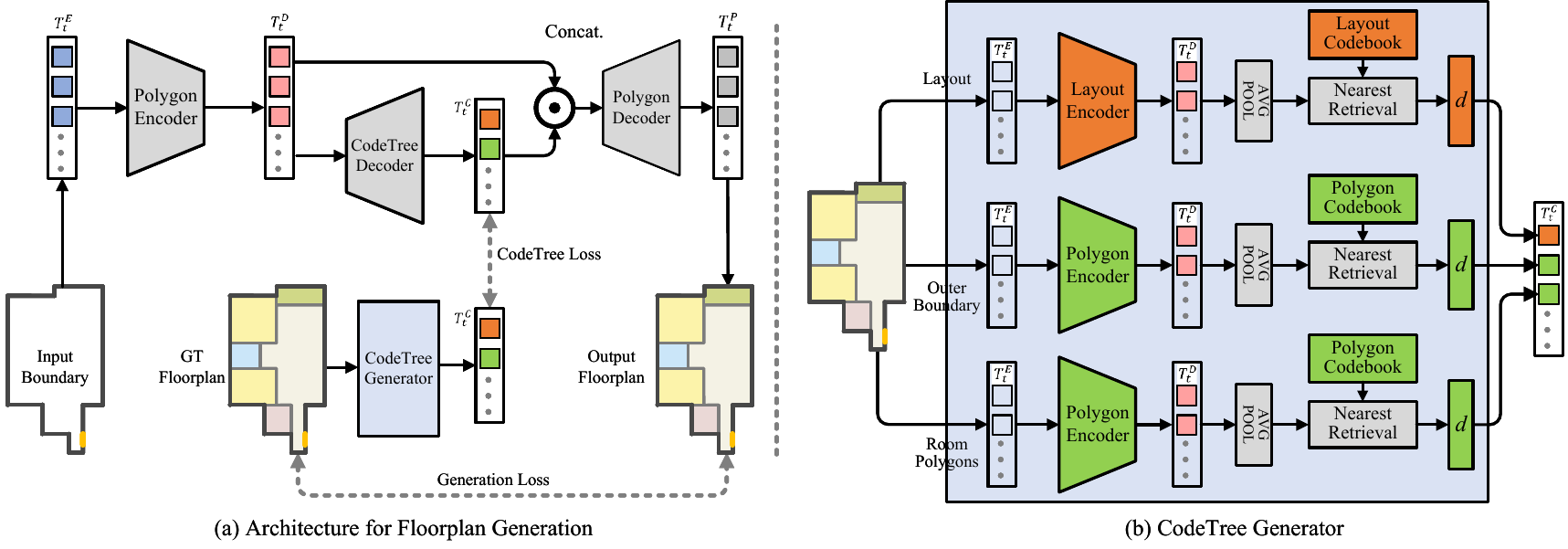}
	\caption{\textbf{CodeTree-based vector floorplan generation:} (a) given an input boundary, the model encodes boundary features, autoregressively predicts a CodeTree, and decodes it into room polygons; (b) the CodeTree Generator encodes the layout, boundary, and room polygons, quantizes them with learned codebooks, and concatenates them into a unified CodeTree for supervision and generation.}
	\label{fig:floorplan_generator}
\end{figure*}

\subsubsection{Network Architecture}

The end-to-end pipeline is illustrated in \Cref{fig:floorplan_generator}(a). During training, a ground-truth floorplan is converted into a supervision CodeTree by tokenizing, encoding, and quantizing the layout, boundary, and room polygons using the pretrained layout and polygon codebooks. In parallel, the \emph{Polygon Encoder} extracts boundary features $T_t^D$, following the same encoding strategy as in polygon codebook learning.

Conditioned on $T_t^D$ and guided by the supervision CodeTree, the \emph{CodeTree Decoder} autoregressively predicts the target CodeTree $T_t^C$. At inference, nucleus (top-$p$) sampling over boundary features $T_t^E$ enables diverse CodeTree generation. The predicted CodeTree $T_t^C$ is concatenated with $T_t^D$ and passed to the \emph{Polygon Decoder}, which generates room polygons $T_t^P$ corner by corner, using delimiter tokens to separate rooms.

To ensure structural consistency across variable-length and irregular floorplans, we fix the room order during training: the living room is placed first, followed by the remaining rooms sorted by their bottom-left coordinates. All components, the Polygon Encoder, CodeTree Decoder, and Polygon Decoder, are implemented using standard Transformer encoder and decoder blocks.

\subsubsection{CodeTree Generator}

The \emph{CodeTree Generator} autoregressively constructs a discrete CodeTree $T_t^C$ that represents the complete floorplan hierarchy, including the global layout, outer boundary, and individual room geometries. Generation is conditioned on boundary features $T_t^E$ and the pretrained layout and polygon codebooks (see \S\ref{sec:codebook}).

As shown in \Cref{fig:floorplan_generator}(b), CodeTree construction proceeds sequentially: the layout code is generated first, followed by the outer boundary (treated as a special polygon), and then the polygon codes for each room. During training, the supervision CodeTree is obtained by tokenizing and quantizing the ground-truth layout and polygons via the learned encoders. This hierarchical representation allows the model to jointly learn global structure and fine-grained geometry directly from the input boundary.

\subsubsection{Training Objective}

CodeTree prediction is formulated as an autoregressive sequence modeling problem. At each decoding step $t$, the model predicts a discrete token $z_t$ from one of three vocabularies: layout or polygon code indices $\mathcal{V}_{\mathrm{code}}$, discretized coordinate bins $\mathcal{V}_{\mathrm{pos}}=\{0,\ldots,63\}$, or room-type labels $\mathcal{V}_{\mathrm{type}}$. Let $\tau_t \in \{\mathrm{code}, \mathrm{pos}, \mathrm{type}\}$ denote the token type at step $t$, and $p_\theta(\cdot \mid \mathbf{z}_{<t})$ the predicted distribution conditioned on the preceding context. The per-step loss is defined by the cross-entropy:
\begin{align}
	\ell_t = - \log p_\theta \left(z_t^\ast \middle| \mathbf{z}_{<t}, \tau_t\right),
\end{align}
where $z_t^\ast$ is the ground-truth token.

We group tokens by type and define the average loss for each group as:
\begin{align}
	\mathcal{L}_{\tau} = \frac{1}{|\mathcal{T}_{\tau}|} \sum_{t \in \mathcal{T}_{\tau}} \ell_t,
	\quad \tau \in \{\mathrm{code}, \mathrm{pos}, \mathrm{type}\}.
\end{align}
The final training objective balances the three components:
\begin{align}
	\mathcal{L} = w_1 \mathcal{L}_{\mathrm{code}}
	+ w_2 \mathcal{L}_{\mathrm{pos}}
	+ w_3 \mathcal{L}_{\mathrm{type}},
\end{align}
where $w_1 = w_2 = w_3 = 1$ in all experiments.

\section{Experimental Results}

\subsection{Experimental Settings}

\subsubsection{Dataset and Evaluation Metrics}

We evaluate TLC-Plan on \textsc{RPLAN} dataset~\cite{wu2019data}, which contains 81{,}235 annotated residential layouts. Each floorplan is stored as a $256\times256$ four-channel image (interior structure, outer boundary, semantic masks, and instance IDs). Following~\cite{zhang2024maskplan}, we split the dataset into train/val/test sets with an 8:1:1 ratio. For vector-based evaluation, we convert raster floorplans into polygonal representations using Graph2Plan~\cite{hu2020graph2plan}, retaining six room types (living room, bedroom, bathroom, kitchen, balcony, and storage).

\begin{figure}[!t]
	\centering
	\includegraphics[width = 0.8\linewidth]{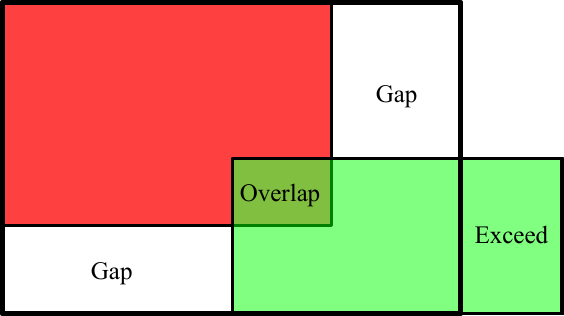}
	\caption{Geometric metrics for vector-based boundary constraints. Given an input boundary polygon (white) and generated room polygons (green and red), we measure \emph{Gap} (uncovered interior area), \emph{Overlap} (intersecting area between rooms), and \emph{Exceed} (area extending beyond the boundary).}
	\label{fig:geometric_metric}
\end{figure}

We report two complementary metric suites. First, we evaluate visual and statistical fidelity using Fr\'echet inception distance ($\mathrm{FID}_{\mathrm{img}}$)~\cite{heusel2017gans} on rendered images, together with mean squared error (MSE) on room counts ($\mathrm{MSE}_T$), adjacency ($\mathrm{MSE}_A$), and sizes ($\mathrm{MSE}_S$)~\cite{zhang2024maskplan}. Second, to measure strict vector consistency under boundary constraints, we define three geometric metrics: Mean Ratio of Gap (MRG), Mean Ratio of Overlap (MRO), and mean ratio of exceed (MRE) (see \Cref{fig:geometric_metric}). For $N$ samples,
\begin{align}
	\mathrm{MRG} & = \frac{1}{N} \sum_{i = 1}^N \frac{A_{\mathrm{gap}}^{(i)}}{A_{\mathrm{boundary}}^{(i)}}, \\
	\mathrm{MRO} & = \frac{1}{N} \sum_{i = 1}^N \frac{A_{\mathrm{overlap}}^{(i)}}{A_{\mathrm{boundary}}^{(i)}}, \\
	\mathrm{MRE} & = \frac{1}{N} \sum_{i = 1}^N \frac{A_{\mathrm{exceed}}^{(i)}}{A_{\mathrm{exceed}}^{(i)} + A_{\mathrm{boundary}}^{(i)}},
\end{align}
where $A_{\mathrm{gap}}^{(i)}$, $A_{\mathrm{overlap}}^{(i)}$, and $A_{\mathrm{exceed}}^{(i)}$ denote the gap, overlap, and exceed areas of the $i$-th sample, and $A_{\mathrm{boundary}}^{(i)}$ is the corresponding boundary area.

\subsubsection{Implementation Details}

All models are implemented in PyTorch with Transformer backbones. The embedding and hidden dimensions are 256, the feedforward dimension is 512, and dropout is 0.1. The codebook learning networks use 4 layers with 8 heads, while the generation network uses 6 layers with 8 heads. The layout and polygon codebooks contain 6{,}000 and 5{,}000 entries, respectively. We cap the number of rooms at 20 and the number of polygon vertices at 40. During codebook training, we randomly mask 30-70\% of tokens. During generation, we cap the CodeTree length at 35 and the polygon sequence length at 800, and apply top-$p$ sampling with $p=0.95$.

\subsubsection{Training Protocol}

All experiments are conducted on a single NVIDIA RTX 4090 GPU. For \textsc{RPLAN} dataset, we apply data augmentation using $90^{\circ}$, $180^{\circ}$, and $270^{\circ}$ rotations together with horizontal and vertical flips, yielding approximately 172{,}000 training samples. The layout and polygon codebook networks are trained independently with a batch size of 512 for 500 epochs, requiring about 3.8 hours and 24.8 hours, respectively. The generation network is trained with a batch size of 256 for 800 epochs, taking roughly 41.6 hours. Coordinates are normalized to $[0,63]$. We optimize all models using AdamW~\cite{loshchilov2017decoupled}, with a 200-step linear warm-up and a peak learning rate of 0.001.

\subsubsection{Baselines}

We compare TLC-Plan with four raster-based methods, RPLAN~\cite{wu2019data}, Graph2Plan~\cite{hu2020graph2plan}, iPLAN~\cite{he2022iplan}, and MaskPLAN~\cite{zhang2024maskplan}, and one vector-based method, GSDiff~\cite{hu2025gsdiff}. RPLAN rasterizes wall lines and heuristically vectorizes them. Graph2Plan retrieves a topology graph from the boundary and uses a CNN-GNN hybrid to generate aligned bounding boxes and rasterized layouts. iPLAN predicts per-pixel room types through cascaded mask refinement, while MaskPLAN employs dynamic masked autoencoders for flexible editing. GSDiff first denoises corner coordinates with diffusion and then infers edges between corner pairs. All baselines rely on post hoc vectorization or separate alignment steps, which can lead to mis-decomposition and misalignment, whereas TLC-Plan operates in a single end-to-end vector pipeline.

\begin{table*}[!t]
	\centering
	\small
	\setlength{\tabcolsep}{9pt}
	\begin{tabular}{c|cr|rc|rccr|ccc}
	\toprule[1.1pt]
	Type & Method & Venue & Size (M) & Time & $\mathrm{FID}_{\mathrm{img}}$ & $\mathrm{MSE}_T$ & $\mathrm{MSE}_A$ & $\mathrm{MSE}_S$ & $\mathrm{MRG}$ & $\mathrm{MRE}$ & $\mathrm{MRO}$ \\
	\midrule
	\multirow{4}{*}{Raster} & RPLAN & TOG'19 & 111 & 1.2 & 9.88 & 0.165 & 1.858 & 24.791 & - & - & - \\
	& Graph2Plan & TOG'20 & \textbf{8} & \textbf{0.1} & 2.36 & \textbf{0.016} & 1.691 & 6.053 & - & - & - \\
	& iPLAN & CVPR'22 & 31 & 2.8 & 3.84 & 0.256 & 2.395 & 18.393 & - & - & - \\
	& MaskPLAN & CVPR'24 & 947 & 2.5 & 10.58 & \underline{0.039} & \textbf{0.058} & 44.420 & - & - & - \\
	\midrule
	\multirow{2}{*}{Vector} & GSDiff & AAAI'25 & 105 & 0.7 & \underline{1.98} & 0.192 & \underline{0.069} & \underline{2.989} & \underline{2.84\%} & \underline{0.45\%} & \textbf{0.00\%} \\
	& \cellcolor{gray!20}TLC-Plan & \cellcolor{gray!20}- & \cellcolor{gray!20}\underline{22} & \cellcolor{gray!20}\underline{0.6} & \cellcolor{gray!20}\textbf{1.84} & \cellcolor{gray!20}0.200 & \cellcolor{gray!20}2.088 & \cellcolor{gray!20}\textbf{2.406} & \cellcolor{gray!20}\textbf{0.71\%} & \cellcolor{gray!20}\textbf{0.10\%} & \cellcolor{gray!20}\underline{0.56\%} \\
	\bottomrule[1.1pt]
	\end{tabular}
	\caption{\textbf{Quantitative comparison of TLC-Plan and state-of-the-art methods on \textsc{RPLAN} dataset.} \textbf{Bold} and \underline{underlined} indicate the best and second-best results, respectively.}
	\label{tab:sota_RPLAN}
\end{table*}

\subsection{Comparisons to State-of-the-Art Methods}

\subsubsection{Quantitative Results}

\Cref{tab:sota_RPLAN} compares TLC-Plan with state-of-the-art methods on \textsc{RPLAN} dataset. Inference time is evaluated by generating one floorplan per image over 3{,}000 samples and averaging the results. TLC-Plan achieves the second-fastest inference speed, surpassed only by Graph2Plan. TLC-Plan attains the best $\mathrm{FID}_{\mathrm{img}}$ of 1.84, indicating superior visual fidelity. For room sizes, TLC-Plan achieves the lowest $\mathrm{MSE}_S$ (2.406), benefiting from accurate polygon-level refinement via two-level code decoding. On room-type counts, TLC-Plan achieves $\mathrm{MSE}_T=0.200$, close to GSDiff (0.192) but higher than Graph2Plan (0.016), reflecting a trade-off between expressive vector synthesis and strict type alignment. For room adjacency, TLC-Plan obtains $\mathrm{MSE}_A=2.088$, comparable to RPLAN (1.858) and iPLAN (2.395) despite not explicitly modeling adjacency graphs; MaskPLAN remains the strongest (0.058) due to dense adjacency supervision.

TLC-Plan also excels in vector-specific geometric consistency, with all three vector metrics below 1\%. It achieves the lowest mean ratio of gap (MRG) at 0.71\% and the lowest mean ratio of exceed (MRE) at 0.10\%, while maintaining a controlled mean ratio of overlap (MRO) at 0.56\%. By comparison, GSDiff exhibits higher MRG (2.84\%) and MRE (0.45\%) but zero MRO, since its corner-based construction inherently avoids overlaps. Overall, these results validate TLC-Plan’s effectiveness in end-to-end vector floorplan generation under geometric and topological constraints.

\begin{figure}[!t]
	\centering
	\includegraphics[width = \linewidth]{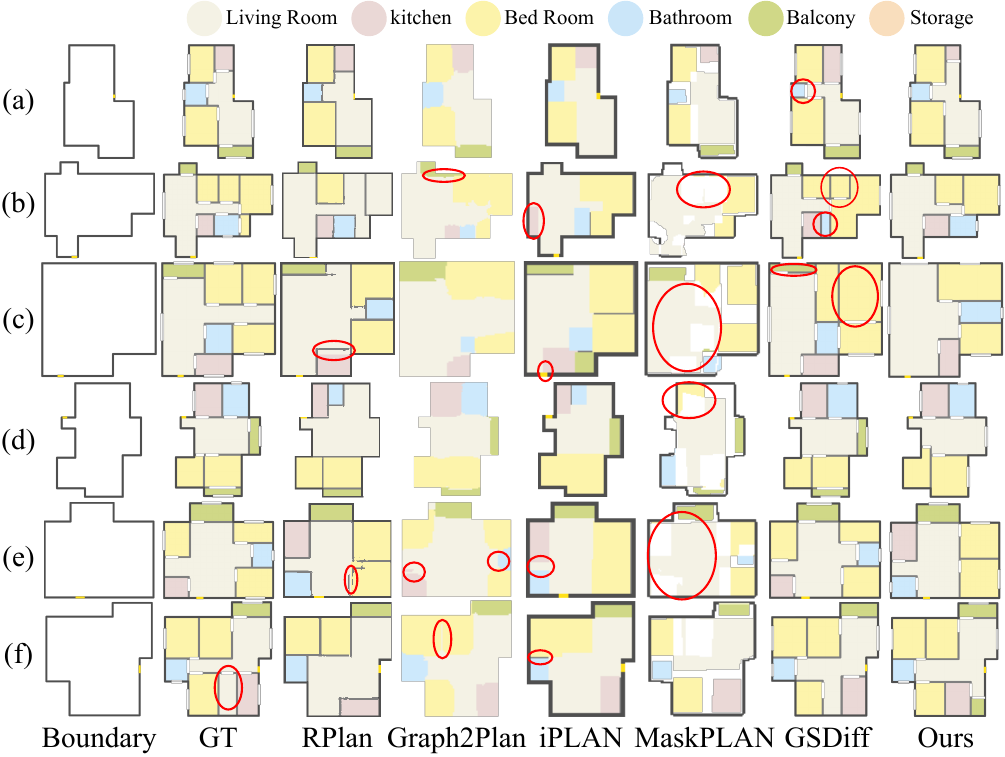}
	\caption{\textbf{Qualitative comparison of floorplan generation:} results from RPLAN, Graph2Plan, iPLAN, MaskPLAN, GSDiff, and our method, with the input boundary and GT shown for reference.}
	\label{fig:sota_rplan}
\end{figure}

\subsubsection{Qualitative Evaluation}

\Cref{fig:sota_rplan} compares floorplans generated by the baselines and TLC-Plan. Raster-based methods (RPLAN, Graph2Plan, iPLAN, MaskPLAN) produce pixel-level outputs that require post-processing, often leading to misalignment, jagged walls, or incomplete rooms. RPLAN frequently yields disjoint interiors and duplicated walls (c, e). Although generally box-aligned, Graph2Plan can generate unrealistic proportions (b, e, f). Without interactive editing, iPLAN may produce door conflicts and area inconsistencies (b, c, e). MaskPLAN, constrained by learned priors, can miss key rooms (b-d) or leave large empty regions (a, e, f). Among vector methods, GSDiff produces plausible layouts (d-f) but provides weaker control over room counts and sizes (a-c). In contrast, TLC-Plan consistently generates well-aligned, coherent, and diverse floorplans that better match the ground truth while satisfying boundary constraints.

\begin{figure}[!t]
	\centering
	\includegraphics[width = 1.0\linewidth]{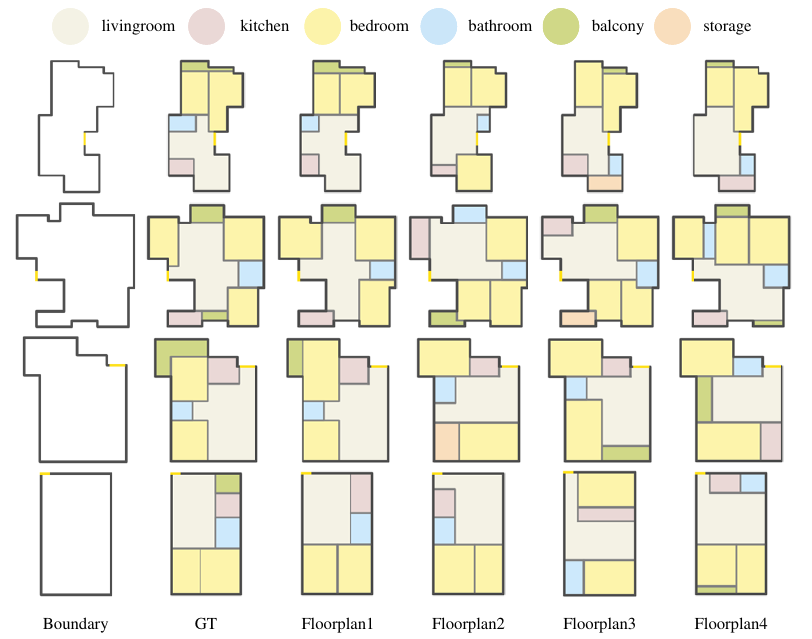}
	\caption{\textbf{Diverse floorplans from a single boundary:} our method generates multiple layouts for the same input boundary, demonstrating sampling variability and design flexibility.}
	\label{fig:diversity_rplan}
\end{figure}

\subsubsection{Layout Diversity}

As shown in \Cref{fig:diversity_rplan}, TLC-Plan generates multiple plausible interior layouts from the same building boundary. The samples vary in room arrangement, orientation, and connectivity, while preserving geometric validity and architectural coherence. This diversity arises from autoregressive sampling over the CodeTree, which encodes both high-level layout patterns and geometric refinements. Unlike deterministic pipelines, TLC-Plan supports design exploration through automatic alternative generation and customization. All sampled outputs remain topologically consistent and semantically complete, demonstrating robust CodeTree-guided generation.

\begin{figure}[!t]
	\centering
	\includegraphics[width = \linewidth]{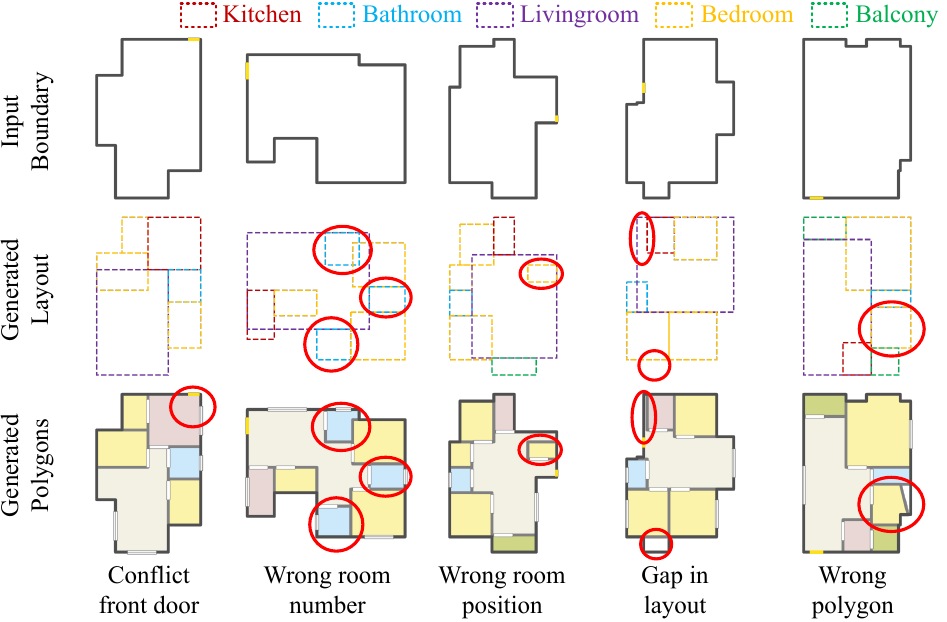}
	\caption{\textbf{Failure cases on \textsc{RPLAN}.}}
	\label{fig:failure_cases}
\end{figure}

\subsubsection{Failure Cases}


While TLC-Plan typically produces high-quality vector floorplans, challenging inputs can expose failure modes (see \Cref{fig:failure_cases}). Common errors on \textsc{RPLAN} include misplaced front doors (\emph{e.g.}, adjacent to kitchens instead of living rooms), unrealistic bathroom counts for small units, bedrooms disconnected from exterior walls, residual boundary gaps, and locally invalid room polygons despite coherent global layouts.


In 3,000 generated samples, front-door misplacement (not adjacent to a living room) occurs in 3.77\% (113) of cases, while fully coherent layouts—without gaps or invalid rooms—are achieved in 86.83\% (2,605) of outputs. These errors primarily stem from three factors. (1) \textbf{Dataset bias:} the \textsc{RPLAN} dataset mainly contains layouts with 2 to 8 rooms, limiting generalization to larger or more complex plans and sometimes resulting in incomplete outputs. (2) \textbf{Codebook quantization:} limited discrete capacity can introduce refinement artifacts during polygon decoding. (3) \textbf{Autoregressive drift:} local sampling errors may accumulate across decoding steps, leading to global inconsistencies such as invalid adjacency or geometry. The model also struggles with highly complex boundaries, such as those with more than 30 corners, which are rare in training data. Some failure cases are further caused by annotation inconsistencies in \textsc{RPLAN} (e.g., case (f) in \Cref{fig:sota_rplan}).

Future improvements include higher-quality training data, expanded codebook capacity, and constrained or guided decoding (e.g., via reinforcement learning) to better enforce architectural validity and functional coherence.

\subsubsection{Extreme Cases}

\begin{figure}[!t]
	\centering
	\includegraphics[width=\linewidth]{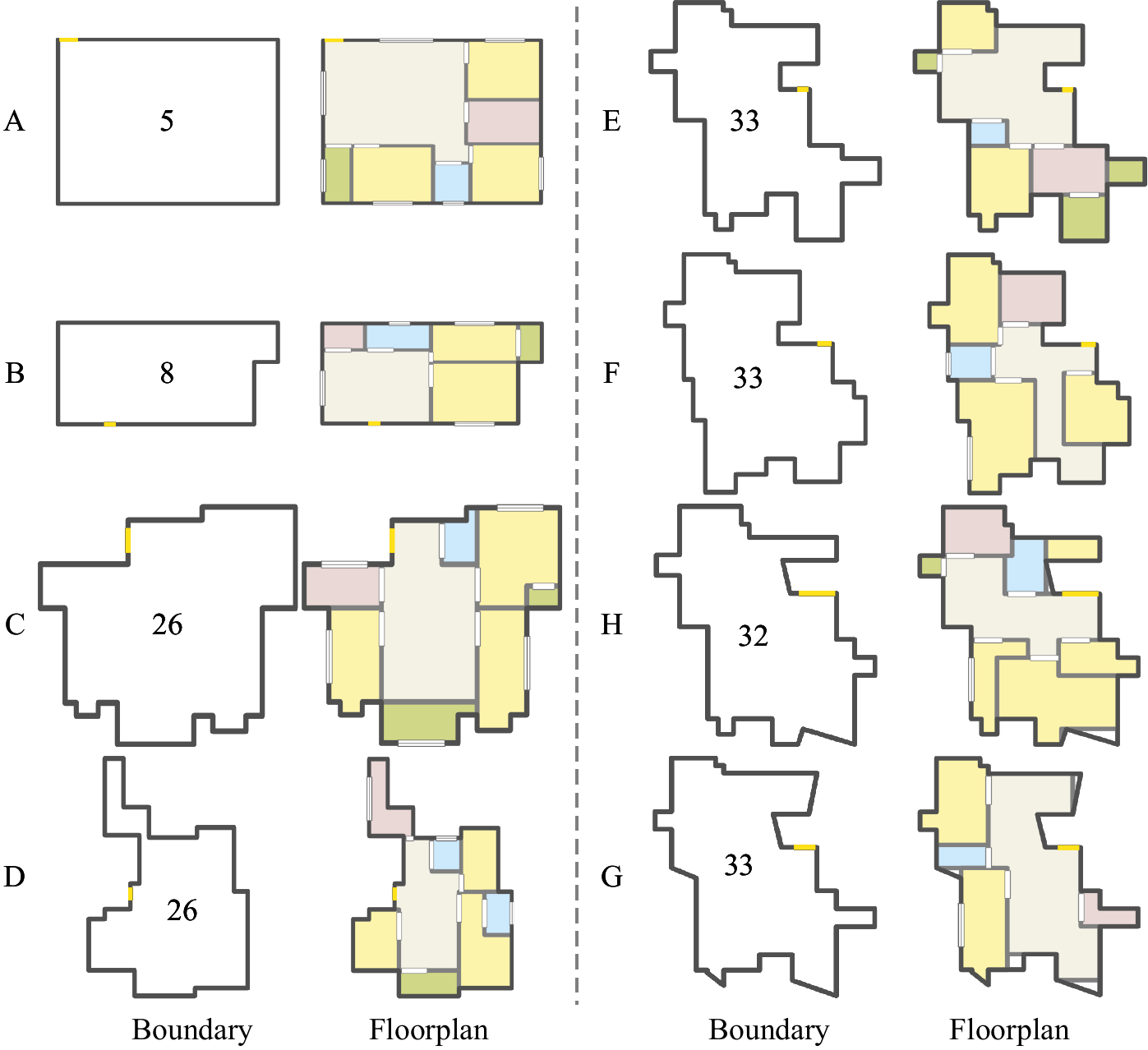}
	\caption{\textbf{Generalization to extreme boundaries.} The numbers denote boundary vertex counts, demonstrating robustness to complex shapes beyond the training distribution.}
	\label{fig:extrem_cases}
\end{figure}

\Cref{fig:extrem_cases} presents floorplans generated from input boundaries that are substantially more complex than those seen during training. In the \textsc{RPLAN} dataset, outer boundaries contain between 4 and 35 vertices, with an average of 7.0, while the majority of training samples fall within 6 to 10 vertices. In contrast, the examples shown here include boundaries with up to 33 vertices, representing a clear distribution shift.

Cases A–F follow the Manhattan assumption and vary widely in boundary complexity, from 5 to 33 vertices. TLC-Plan generates coherent and geometrically valid layouts across all these cases, demonstrating robustness to increased boundary complexity. Cases G and H further evaluate generalization beyond the training distribution by introducing slanted boundary edges, which are not present in \textsc{RPLAN} dataset. Although small gaps appear near the slanted edges, the overall layouts remain reasonable and structurally consistent. These results indicate that TLC-Plan generalizes well to extreme and partially out-of-distribution boundary conditions, while also highlighting limitations when assumptions in the training data are violated.

\begin{figure}[!t]
	\centering
	\includegraphics[width = \linewidth]{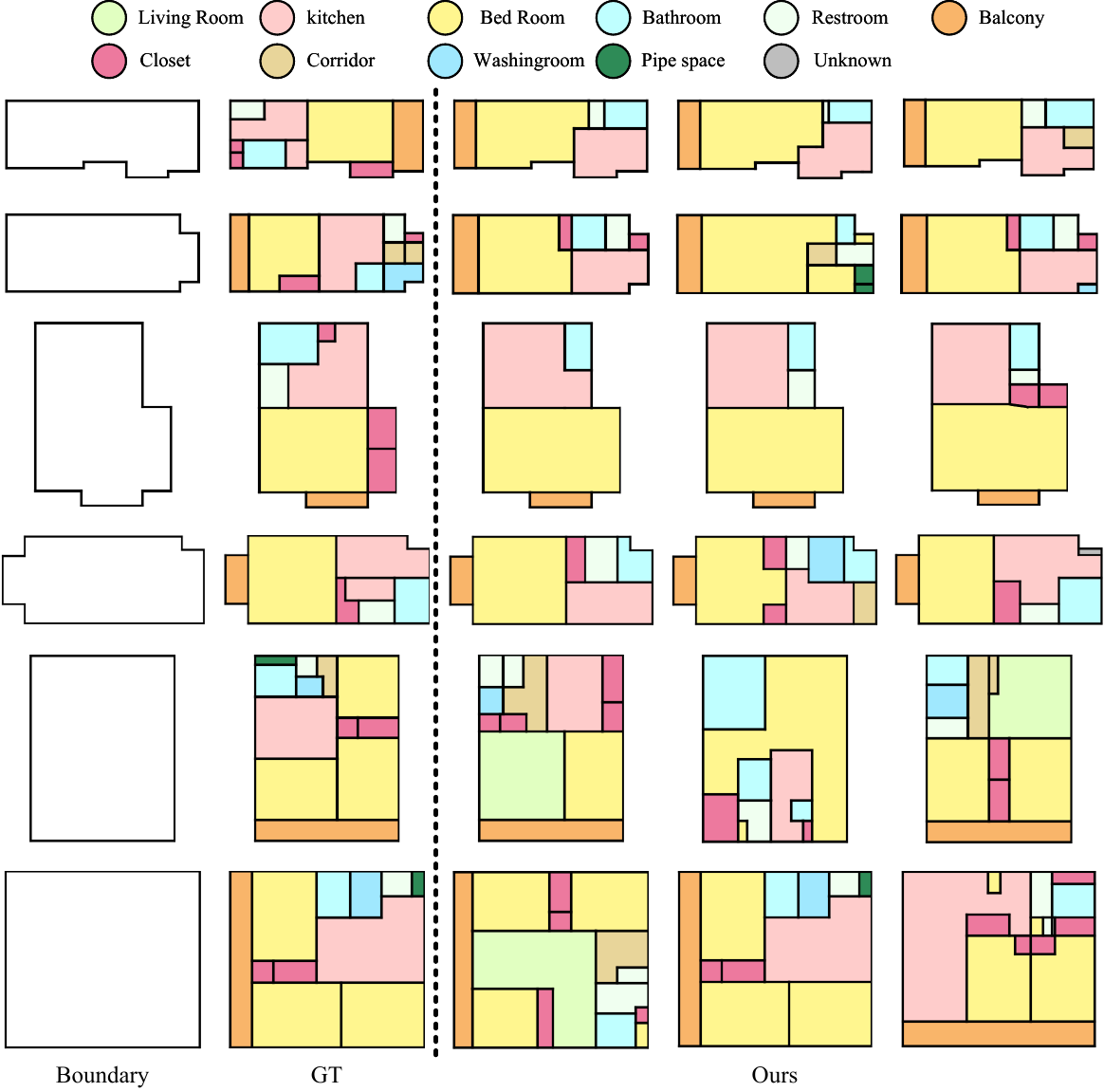}
	\caption{\textbf{Qualitative evaluation on \textsc{LIFULL}.}}
	\label{fig:sota_lifull}
\end{figure}

\begin{table}[!t]
	\centering
	\small
	\setlength{\tabcolsep}{12pt}
	\begin{tabular}{c|rrrr}
	\toprule[1.1pt]
	Method & $\mathrm{FID}_{\mathrm{img}}$ & $\mathrm{MSE}_T$ & $\mathrm{MSE}_A$ & $\mathrm{MSE}_S$ \\
	\midrule
	RPLAN & 50.19 & 5.15 & - & 159.81 \\
	iPLAN & 37.35 & 2.52 & - & 36.55 \\
	GSDiff & 12.44 & 0.98 & \textbf{0.14} & 20.12 \\
	\rowcolor{gray!20}
	Our & \textbf{9.08} & \textbf{0.34} & 4.86 & \textbf{8.96} \\
	\bottomrule[1.1pt]
	\end{tabular}
	\caption{\textbf{Quantitative evaluation on \textsc{LIFULL} dataset.}}
	\label{tab:sota_lifull}
\end{table}

\subsubsection{Generalization to \textsc{LIFULL}}\label{sec:lifull_eval}

We evaluate generalization on \textsc{LIFULL}~\cite{lifull2016}, which contains 10{,}804 Japanese residential floorplans vectorized via Raster-to-Graph~\cite{hu2024raster}. Using the same training configuration as for \textsc{RPLAN}, we only adjust the room taxonomy to match \textsc{LIFULL}'s 12-class scheme. Baseline results for RPLAN and iPLAN~\cite{he2022iplan} (without $\mathrm{MSE}_A$), and for GSDiff~\cite{hu2025gsdiff}. As shown in \Cref{tab:sota_lifull,fig:sota_lifull}, TLC-Plan achieves the best $\mathrm{FID}_{\mathrm{img}}$ (9.08), $\mathrm{MSE}_T$ (0.34), and $\mathrm{MSE}_S$ (8.96), demonstrating strong visual realism, type accuracy, and size consistency. While GSDiff performs better on room adjacency ($\mathrm{MSE}_A$), it lags in overall fidelity. These results confirm TLC-Plan’s ability to generalize across datasets with distinct distributions. However, as noted in~\cite{hu2024raster}, the lower annotation quality of \textsc{LIFULL} can limit vectorization fidelity and evaluation reliability.

\subsection{Ablation Study}

We conduct ablation studies to assess the impact of type encoding, architectural components, masking strategies, and codebook sizes. These experiments isolate the contribution of each factor and validate our design choices. Overall, the masked skip connection markedly improves generalization, while properly sized codebooks balance representational capacity and training stability.

Based on these findings, we adopt the following configuration in all experiments (see \S\ref{sec:method}): \emph{(i)} omit type encoding in the polygon layer to improve geometric generalization; \emph{(ii)} retain all architectural components, including the masked skip connection and hierarchical codetree; \emph{(iii)} apply a random masking ratio of 30\%-70\% during VQ-VAE training; \emph{(iv)} set codebook sizes to 6{,}000 for layouts and 5{,}000 for polygons; and \emph{(v)} discretize layout tuples at 6-bit resolution.

\subsubsection{Type Encoding in the Polygon Layer}\label{sec:label_encode}

In the layout layer, room types are encoded in $(x, y, w, h, c)$, where $c$ denotes the semantic class. At the polygon level, we examine whether type labels benefit geometric modeling. Counterintuitively, removing type information yields slightly better performance, likely because semantics are already conveyed by the codetree and layout codebook, allowing the polygon codebook to focus on geometry.

\begin{figure}[!t]
	\centering
	\includegraphics[width = 0.8\linewidth]{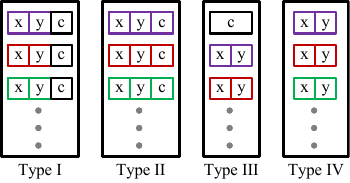}
	\caption{\textbf{Polygon type encoding:} (I) uniform room-type label on each vertex; (II) mixed labels with a dedicated front-door tag; (III) single prefix label for the entire polygon; (IV) no type labels (coordinates only).}
	\label{fig:label_encode}
\end{figure}

\begin{table}[!t]
	\centering
	\small
	\setlength{\tabcolsep}{4pt}
	\begin{tabular}{c|ccccccc}
	\toprule[1.1pt]
	Type & $\mathrm{FID}_{\mathrm{img}}$ & $\mathrm{MSE}_T$ & $\mathrm{MSE}_A$ & $\mathrm{MSE}_S$ & $\mathrm{MRG}$ & $\mathrm{MRE}$ & $\mathrm{MRO}$ \\
	\midrule
	I & 1.91 & 0.227 & 2.201 & 2.620 & 1.40\% & 0.11\% & 1.12\% \\
	II & 1.96 & 0.228 & 2.193 & 2.555 & 0.90\% & 0.15\% & 1.14\% \\
	III & 1.93 & 0.221 & 2.170 & 2.435 & 1.07\% & 0.18\% & 0.94\% \\
	\rowcolor{gray!20}
	IV & \textbf{1.84} & \textbf{0.200} & \textbf{2.088} & \textbf{2.406} & \textbf{0.71}\% & \textbf{0.10}\% & \textbf{0.56}\% \\
	\bottomrule[1.1pt]
	\end{tabular}
	\caption{\textbf{Effect of different polygon type encoding strategies.}}
	\label{tab:label_encode}
\end{table}

We compare four strategies (see \Cref{fig:label_encode}): Type~I assigns a uniform type label to all vertices; Type~II additionally marks front-door corners; Type~III prepends a single type label to the vertex sequence; and Type~IV uses coordinates only. As shown in \Cref{tab:label_encode}, Type~IV consistently performs best across all metrics, supporting the benefit of decoupling semantics from geometry.

\begin{table}[!t]
	\centering
	\small
	\setlength{\tabcolsep}{9pt}
	\begin{tabular}{l|rrrr}
	\toprule[1.1pt]
	Structure & $\mathrm{FID}_{\mathrm{img}}$ & $\mathrm{MSE}_T$ & $\mathrm{MSE}_A$ & $\mathrm{MSE}_S$ \\
	\midrule
	\emph{w/o} Masked Skip & 16.49 & 2.107 & 5.117 & 8.695 \\
	\emph{w/o} CodeTree & 26.97 & 1.950 & 8.132 & 14.760 \\
	\emph{w/o} Layout Code & 24.29 & 1.282 & 6.977 & 12.553 \\
	\emph{w/o} Polygon Code & 14.01 & 1.020 & 5.244 & 6.248 \\
	\rowcolor{gray!20}
	Full Method & \textbf{1.84} & \textbf{0.200} & \textbf{2.088} & \textbf{2.406} \\
	\bottomrule[1.1pt]
	\end{tabular}
	\caption{\textbf{Ablation study of key network components.}}
	\label{tab:network_structure}
\end{table}

\subsubsection{Component Effect}

\Cref{tab:network_structure} compares five variants: removing the masked skip connection (\emph{w/o} Masked Skip), removing the codetree (\emph{w/o} CodeTree), using only layout codes (\emph{w/o} Polygon Code), using only polygon codes (\emph{w/o} Layout Code), and the full model. Removing the masked skip connection significantly degrades performance, confirming its role in preventing token memorization and encouraging abstraction. Eliminating the codetree yields the worst results, as the model loses intermediate layout guidance. Using only one codebook sacrifices either global structure (polygon-only) or fine geometry (layout-only). The full model consistently performs best, indicating that both codebooks and the codetree are essential for structured vector floorplan generation.

\begin{table}[!t]
	\centering
	\small
	\setlength{\tabcolsep}{10pt}
	\begin{tabular}{c|cccc}
	\toprule[1.1pt]
	Masking Range & $\mathrm{FID}_{\mathrm{img}}$ & $\mathrm{MSE}_T$ & $\mathrm{MSE}_A$ & $\mathrm{MSE}_S$ \\
	\midrule
	10\%-90\% & 1.875 & 0.202 & \textbf{2.082} & 2.407 \\
	20\%-80\% & 1.866 & 0.201 & 2.085 & 2.426 \\
	\rowcolor{gray!20}
	30\%-70\% & \textbf{1.836} & \textbf{0.200} & 2.088 & \textbf{2.406} \\
	40\%-60\% & 1.857 & 0.201 & 2.086 & 2.418 \\
	\bottomrule[1.1pt]
	\end{tabular}
	\caption{\textbf{Effect of masking ratio on generation quality.}}
	\label{tab:mask_ratio2}
\end{table}

\subsubsection{Mask Ratio}

To improve codebook generalization and promote pattern reuse, we train VQ-VAEs with a masked skip connection: a random portion of input tokens is masked and reconstructed from the latent codes. The masking ratio is sampled uniformly from intervals centered at 50\%. As shown in \Cref{tab:mask_ratio2}, the 30\%-70\% range yields the best overall results, achieving the lowest $\mathrm{FID}_{\mathrm{img}}$, $\mathrm{MSE}_T$, and competitive $\mathrm{MSE}_S$. Wider ranges introduce excessive variance, while narrower ranges reduce reconstruction difficulty. We therefore use 30\%-70\% masking in all experiments.

\begin{table}[!t]
	\centering
	\small
	\setlength{\tabcolsep}{5pt}
	\begin{tabular}{c|c|cccc}
	\toprule[1.1pt]
	\begin{tabular}{c}Layout \\ Codebook \\ Size\end{tabular} &
	\begin{tabular}{c}Polygon \\ Codebook \\ Size\end{tabular} &
	$\mathrm{FID}_{\mathrm{img}}$ & $\mathrm{MSE}_T$ & $\mathrm{MSE}_A$ & $\mathrm{MSE}_S$ \\
	\midrule
	4{,}000 & \multirow{5}[8]{*}{5{,}000} & 3.26 & 0.373 & 3.151 & 3.586 \\
	\cmidrule(lr){1-1}\cmidrule(lr){3-6}
	5{,}000 & & 2.68 & 0.262 & 2.467 & 3.054 \\
	\cmidrule(lr){1-1}\cmidrule(lr){3-6}
	\textbf{6{,}000} & & \textbf{1.84} & \textbf{0.200} & \textbf{2.088} & \textbf{2.406} \\
	\cmidrule(lr){1-1}\cmidrule(lr){3-6}
	7{,}000 & & 1.86 & 0.202 & 2.106 & 2.443 \\
	\cmidrule(lr){1-1}\cmidrule(lr){3-6}
	8{,}000 & & 1.90 & 0.218 & 2.173 & 2.598 \\
	\midrule
	\multirow{5}[8]{*}{6{,}000} & 3{,}000 & 2.89 & 0.223 & 3.444 & 4.255 \\
	\cmidrule(lr){2-6}
	& 4{,}000 & 2.07 & 0.212 & 2.655 & 3.098 \\
	\cmidrule(lr){2-6}
	& \textbf{5{,}000} & \textbf{1.84} & \textbf{0.200} & \textbf{2.088} & \textbf{2.406} \\
	\cmidrule(lr){2-6}
	& 6{,}000 & 1.92 & 0.204 & 2.135 & 2.779 \\
	\cmidrule(lr){2-6}
	& 7{,}000 & 2.02 & 0.212 & 2.303 & 2.953 \\
	\bottomrule[1.1pt]
	\end{tabular}
	\caption{\textbf{Performance under different layout and polygon codebook sizes.}}
	\label{tab:cb_size}
\end{table}

\subsubsection{Codebook Size}

Codebook size affects both generation quality and stability. Larger codebooks can represent finer-grained patterns but may reduce utilization and impair generalization, while smaller codebooks are efficient but risk insufficient expressiveness. For the layout codebook, increasing the size from 4{,}000 to 6{,}000 consistently improves all metrics (see \Cref{tab:cb_size}), suggesting that smaller codebooks cannot adequately separate distinct layout patterns. Further increasing beyond 6{,}000 yields diminishing returns and slight degradation, likely due to over-parameterization and lower effective usage. We therefore set the layout codebook size to 6{,}000.

For the polygon codebook, performance improves substantially from 3{,}000 to 5{,}000 entries, indicating that small codebooks cannot capture geometric variability. Beyond 5{,}000, gains saturate and slightly drop, again suggesting reduced generalization with overly large codebooks. We thus fix the polygon codebook size at 5{,}000, which provides the best trade-off between expressiveness and stability.

\begin{table}[!t]
	\centering
	\small
	\setlength{\tabcolsep}{12pt}
	\begin{tabular}{l|rrrr}
	\toprule[1.1pt]
	Bits & $\mathrm{FID}_{\mathrm{img}}$ & $\mathrm{MSE}_T$ & $\mathrm{MSE}_A$ & $\mathrm{MSE}_S$ \\
	\midrule
	5-bit & 2.73 & 0.190 & 2.081 & 2.529 \\
	\rowcolor{gray!20}
	6-bit & 1.84 & 0.200 & 2.088 & 2.406 \\
	7-bit & 2.81 & 0.186 & 2.048 & 3.026 \\
	\bottomrule[1.1pt]
	\end{tabular}
	\caption{\textbf{Effect of layout tuple quantization bit width.}}
	\label{tab:network_structure}
\end{table}

\subsubsection{Limits on Layout Size and Complexity}

The ablation on quantization bit width reveals a clear trade-off between representational precision and layout complexity. As shown in \Cref{tab:network_structure}, 6-bit quantization achieves the best balance, yielding strong visual fidelity ($\mathrm{FID}_{\mathrm{img}}=1.84$) and room-size accuracy ($\mathrm{MSE}_S=2.406$). In contrast, both 5-bit and 7-bit settings lead to degraded performance. With 5-bit quantization, insufficient resolution limits geometric detail, resulting in oversimplified contours. With 7-bit quantization, the increased discrete space introduces redundant variability under a fixed model capacity, which destabilizes long-range geometric dependency modeling during autoregressive decoding and degrades scale consistency, as reflected by the higher $\mathrm{MSE}_S$ (3.026).

Varying the quantization bit width has only a minor impact on topological adjacency error ($\mathrm{MSE}_A$) and functional semantics ($\mathrm{MSE}_T$), indicating that symbolic layout structure and continuous geometry are handled by different components of the model. Overall, quantization acts as a bottleneck that limits the amount of geometric information that can be compressed and faithfully reconstructed. As layout irregularity and functional complexity increase, the fixed-size codebook and autoregressive decoding struggle to jointly preserve fine-scale geometry and global structure. Within the complexity range of the target data, 6-bit quantization therefore represents a practical local optimum. Scaling to larger and more complex layouts will likely require architectural extensions, such as adaptive codebooks or hybrid representations that decouple global topology from local geometry to better allocate representational capacity.

\section{Conclusions and Limitations}

TLC-Plan introduces a hierarchical VQ-VAE framework for end-to-end vector floorplan generation that synthesizes layouts directly from input boundaries and avoids rasterization. By encoding global layout and local geometry into a unified codetree, it reflects human design workflows and supports compositional generalization. Experiments on \textsc{RPLAN} and \textsc{LIFULL} demonstrate state-of-the-art performance in structural coherence, geometric accuracy, and layout diversity, providing a scalable foundation for AI-driven architectural design. More broadly, TLC-Plan moves toward constraint-aware spatial design agents that unify hierarchical planning and geometric reasoning for architectural and urban applications.

Despite these strengths, TLC-Plan currently conditions only on the boundary and focuses on room partitioning. Future work will incorporate richer constraints, such as textual prompts, fixed structural elements, or predefined adjacencies. The codetree formulation is flexible and can integrate such conditions by conditioning autoregressive decoding on additional constraint tokens or masking subsequences to enforce fixed placements. Further directions include modeling interior elements (\emph{e.g.}, doors, windows, furniture) and scaling to multi-floor buildings and city-scale layouts, enabling more complex automated design scenarios.



\bibliographystyle{eg-alpha-doi}
\bibliography{main}


\clearpage
\appendix
\onecolumn
\setcounter{section}{0}
\renewcommand{\thesection}{\Alph{section}}

\begin{center}
\Large\bfseries Supplementary Materials\par
\end{center}\vspace{0.5\baselineskip}

\section{Additional Results and Visualizations}
\FloatBarrier

\begin{figure}[H]
  \centering
  \includegraphics[width=0.75\linewidth]{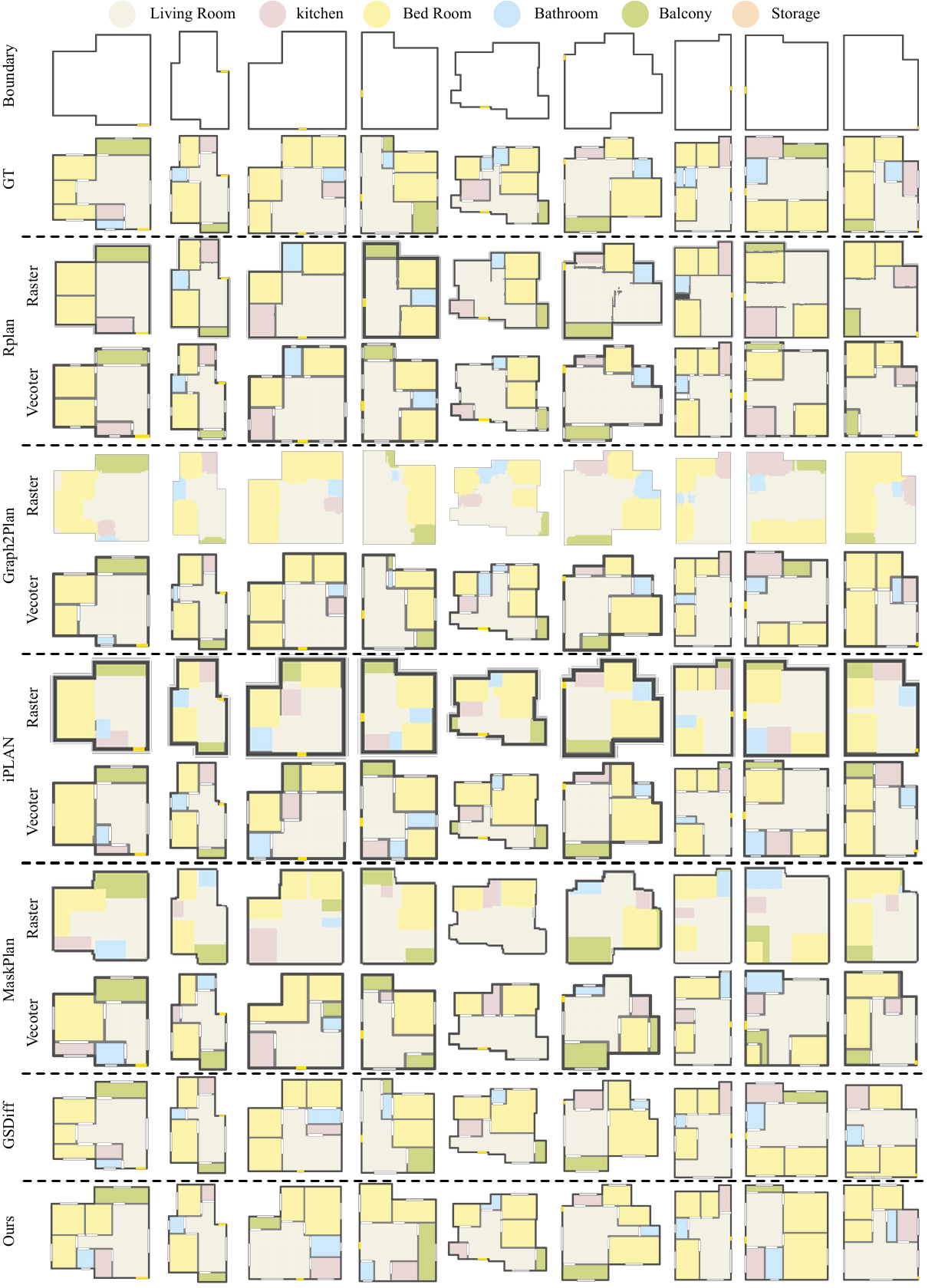}
  \caption{\textbf{Additional qualitative comparison on \textbf{RPLAN} (Part I).}}
  \label{fig:raster_vecotr}
\end{figure}

\begin{figure}[H]
  \centering
  \includegraphics[width=0.80\linewidth]{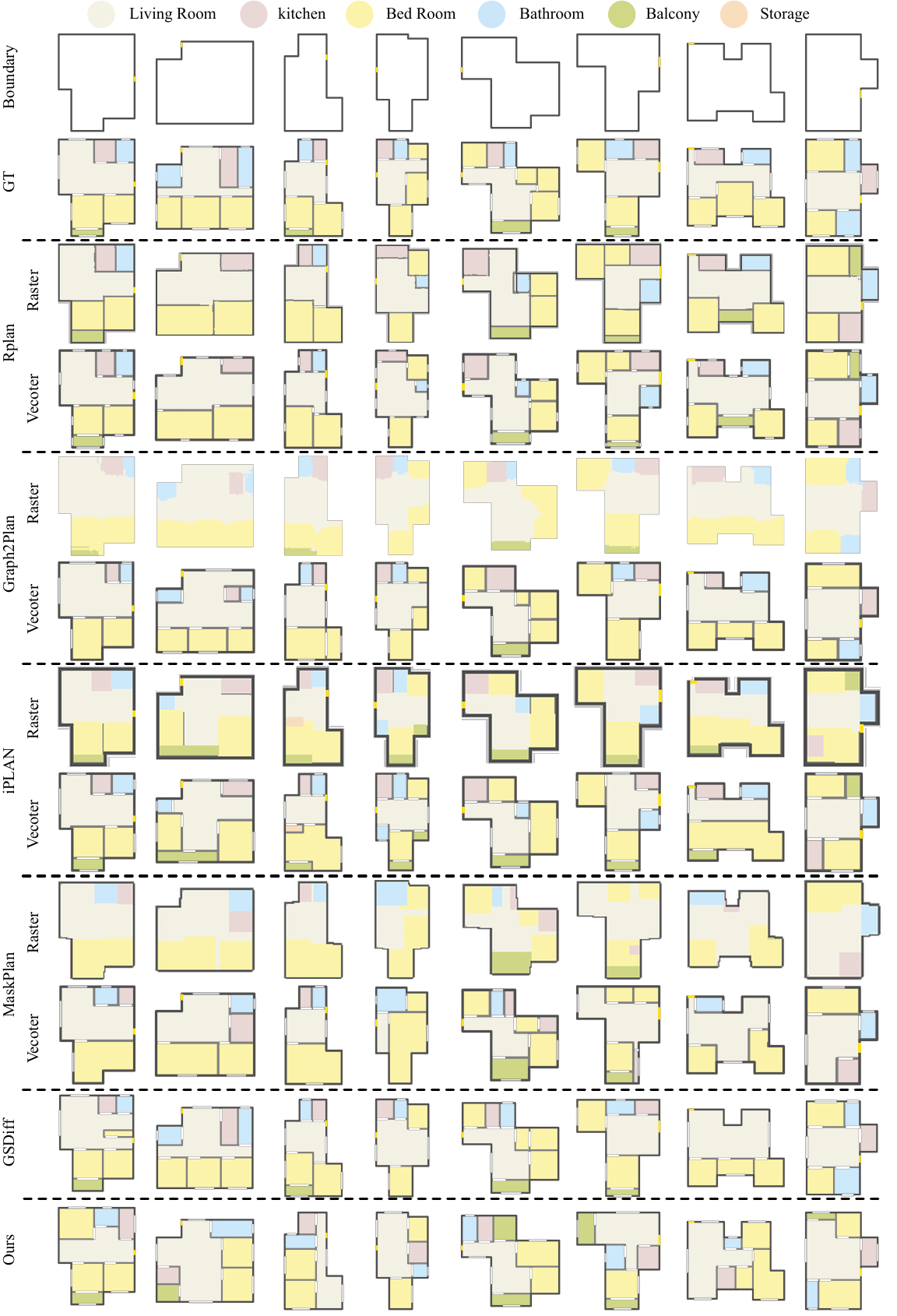}
  \caption{\textbf{Additional qualitative comparison on \textbf{RPLAN} (Part II).}}
  \label{fig:raster_vecotr2}
\end{figure}

\begin{figure}[H]
  \centering
  \includegraphics[width=\linewidth]{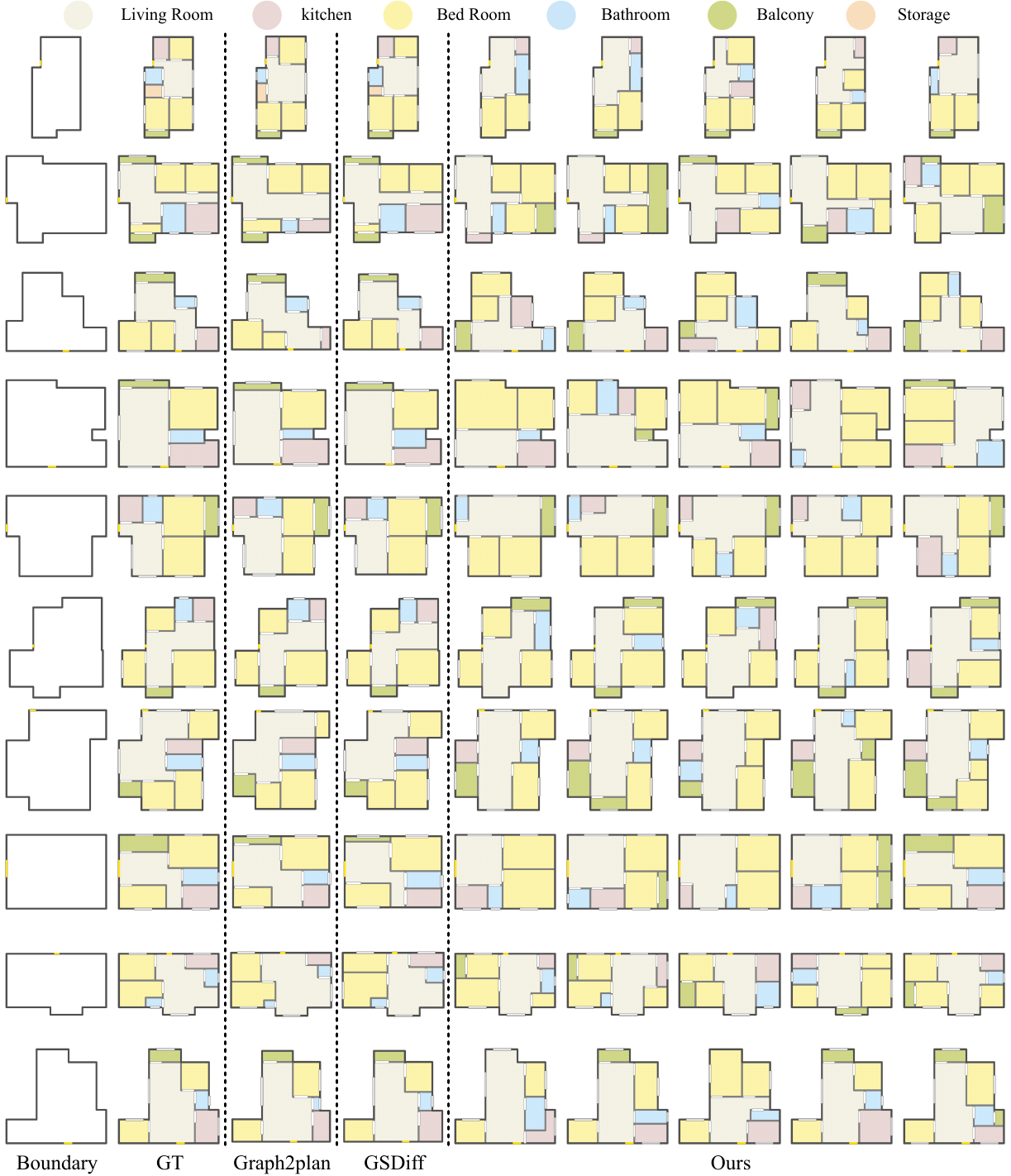}
  \caption{\textbf{Diverse generations from a single boundary on \textbf{RPLAN} (Part I).}}
  \label{fig:sota_rplan_sup1}
\end{figure}

\begin{figure}[H]
  \centering
  \includegraphics[width=\linewidth]{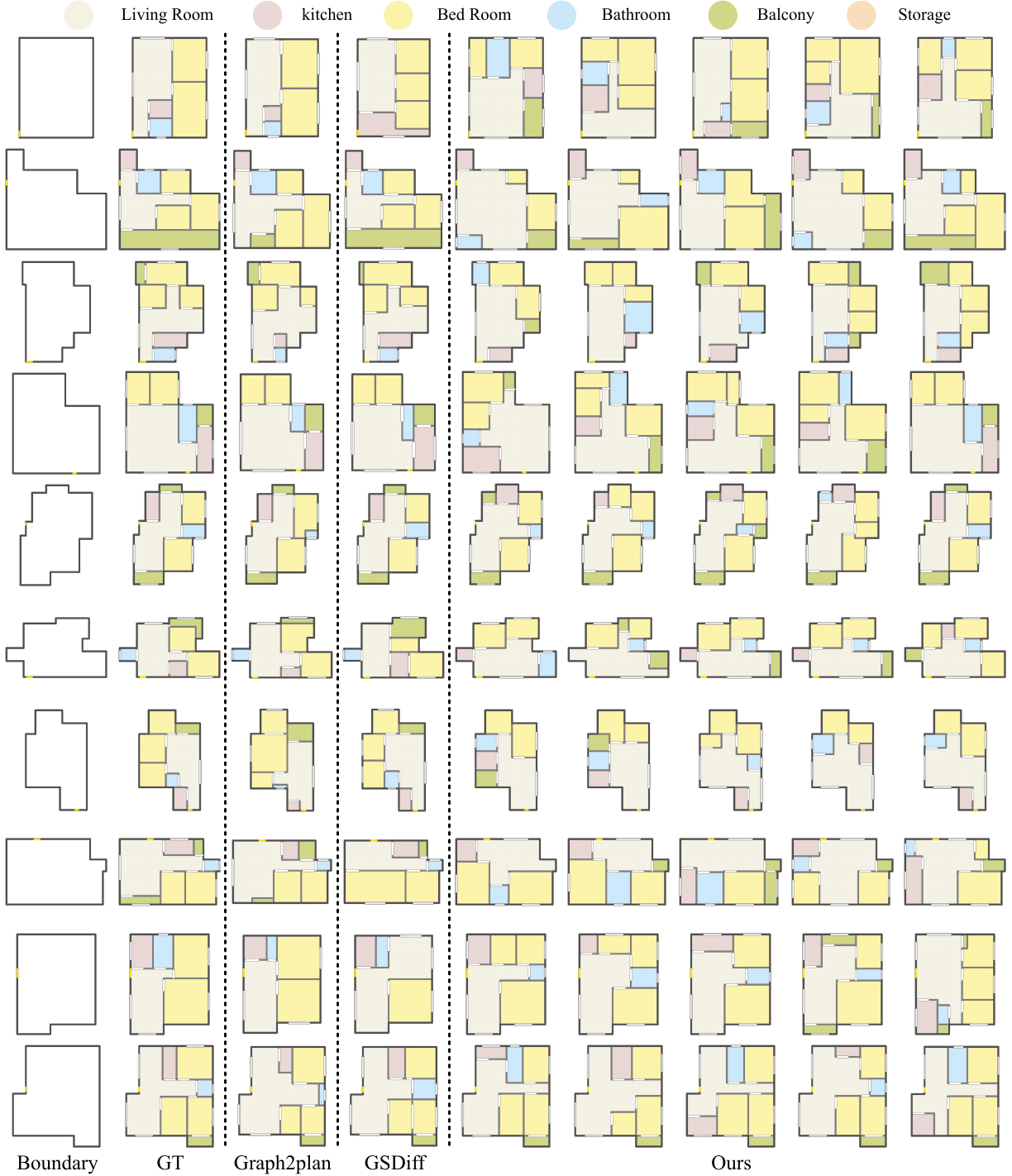}
  \caption{\textbf{Diverse generations from a single boundary on \textbf{RPLAN} (Part II).}}
  \label{fig:sota_rplan_sup2}
\end{figure}

\end{document}